\documentclass[final]{cvpr}

\usepackage{times}
\usepackage{epsfig}
\usepackage{graphicx}
\usepackage{amsmath}
\usepackage{amssymb}

\usepackage{multirow}
\usepackage{caption}
\usepackage{subcaption}
\usepackage{booktabs}
\usepackage{bbm}

\usepackage{color}
\usepackage{balance}

\newcommand{\argmin}{\mathop{\mathrm{argmin\,}}}

\newcommand{\methodname}{Faster Meta Update Strategy}
\newcommand{\methodnameabbre}{FaMUS}

\usepackage[pagebackref=true,breaklinks=true,colorlinks,bookmarks=false]{hyperref}

\begin{document}

\title{Faster Meta Update Strategy for Noise-Robust Deep Learning}

\author{Youjiang Xu$^1$ \hspace{12pt} Linchao Zhu$^2$ \hspace{12pt} Lu Jiang$^3$ \hspace{12pt}  Yi Yang$^2$\\
$^1$Baidu Research $^2$ReLER, University of Technology Sydney $^3$Google Research\\
{\tt\small youjiangxu@gmail.com, lujiang@google.com, \{linchao.zhu, yi.yang\}@uts.edu.au}
}

\maketitle
\thispagestyle{empty} 


\begin{abstract}
It has been shown that deep neural networks are prone to overfitting on biased training data. Towards addressing this issue, meta-learning employs a meta model for correcting the training bias. 
Despite the promising performances, super slow training is currently the bottleneck in the meta learning approaches.
In this paper, we introduce a novel \methodname\ (\methodnameabbre) to replace the most expensive step in the meta gradient computation with a faster layer-wise approximation. We empirically find that \methodnameabbre\xspace yields not only a reasonably accurate but also a low-variance approximation of the meta gradient.
We conduct extensive experiments to verify the proposed method on two tasks. We show our method is able to save two-thirds of the training time while still maintaining the comparable or achieving even better generalization performance. In particular, our method achieves the state-of-the-art performance on
both synthetic and realistic noisy labels, and obtains promising performance on long-tailed recognition on standard benchmarks. Code are released at {\small \url{https://github.com/youjiangxu/FaMUS}}.

\end{abstract}

\begin{figure}[th]
\begin{subfigure}{.45\textwidth}
\centering
\includegraphics[width=0.95\linewidth]{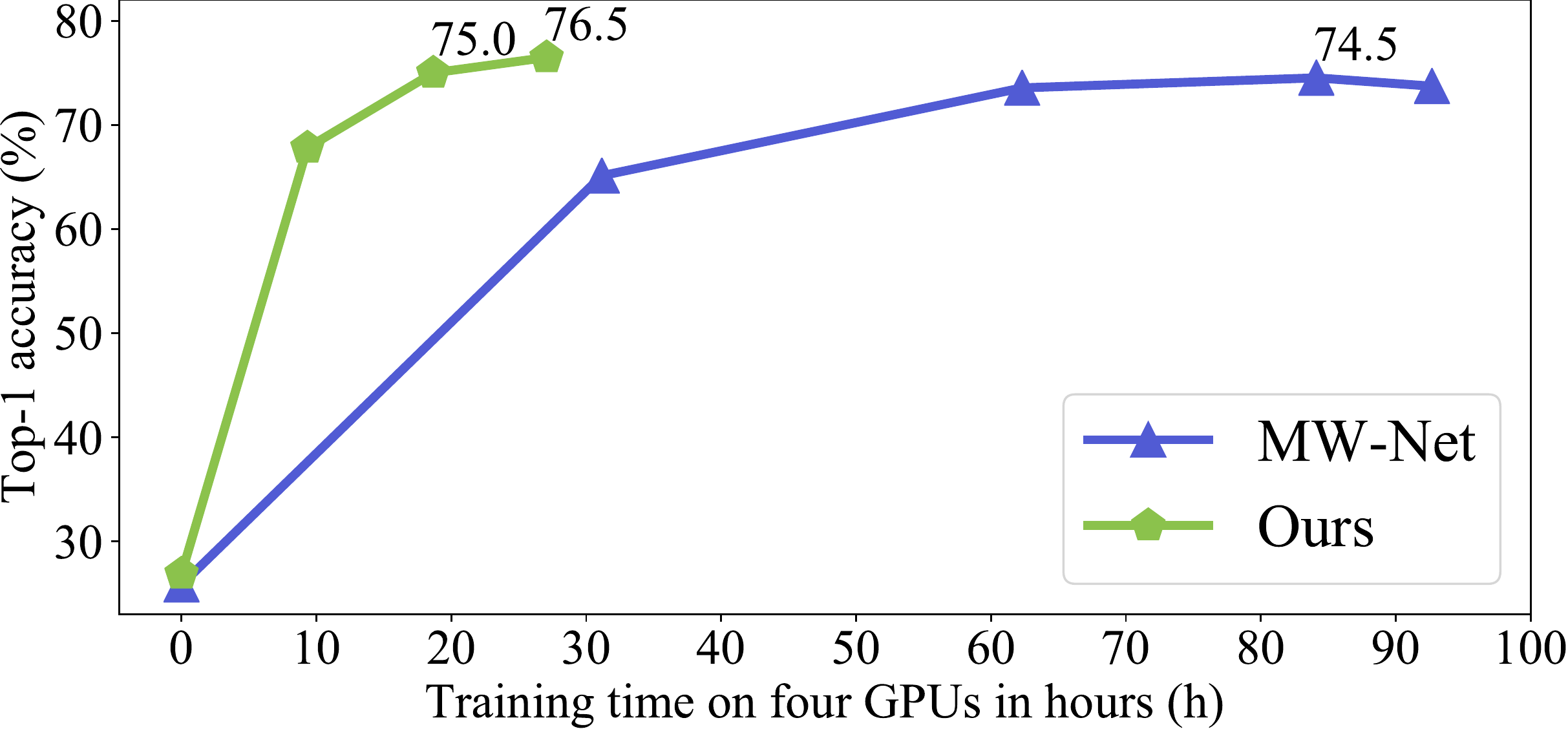}
\footnotesize\caption{Top-1 Accuracy vs. Training time}
\end{subfigure}
\begin{subfigure}{.23\textwidth}
\includegraphics[width=1.0\linewidth]{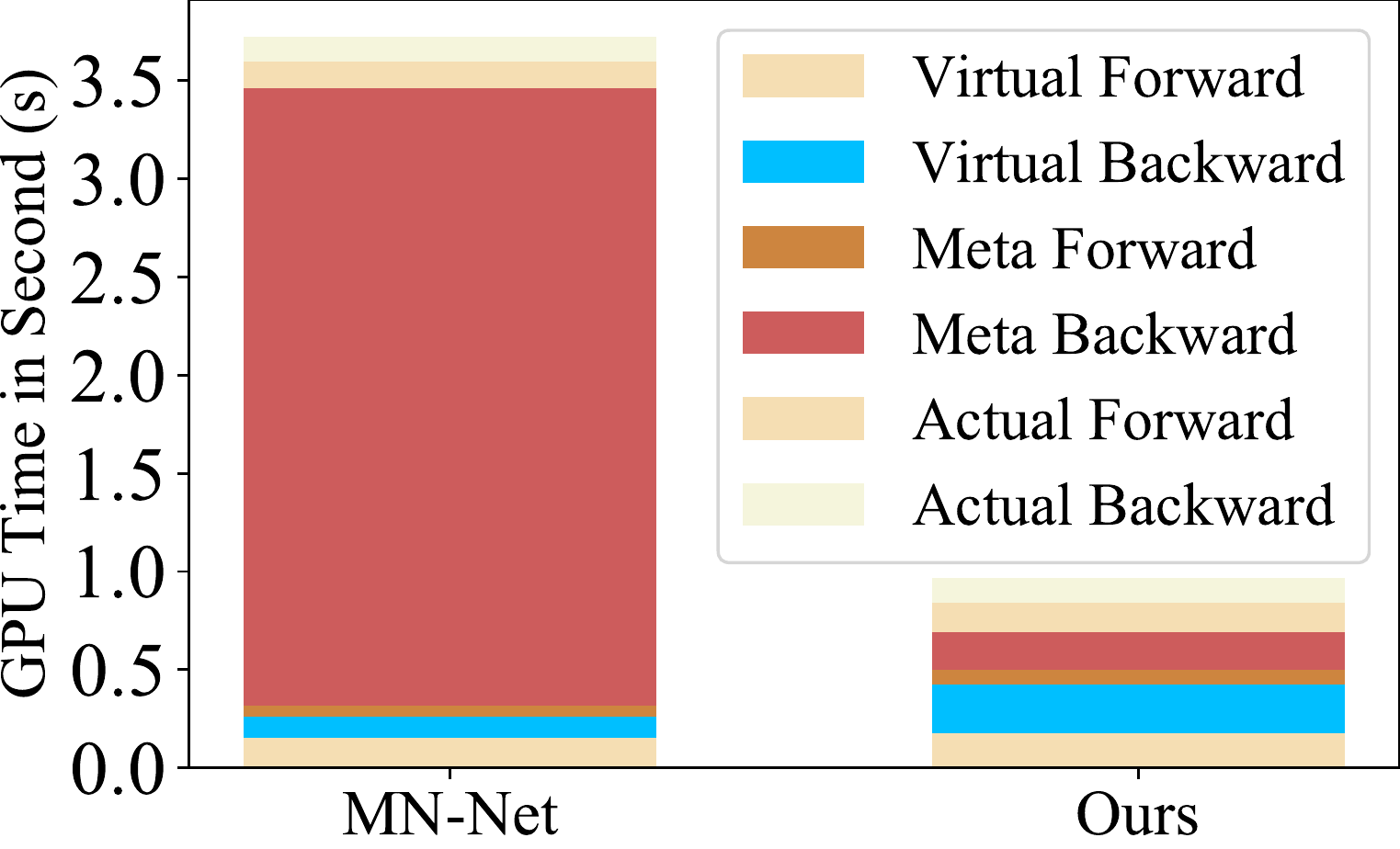}
\footnotesize\caption{Cost in One Iteration}
\end{subfigure}
\begin{subfigure}{.23\textwidth}
\includegraphics[width=1.0\linewidth]{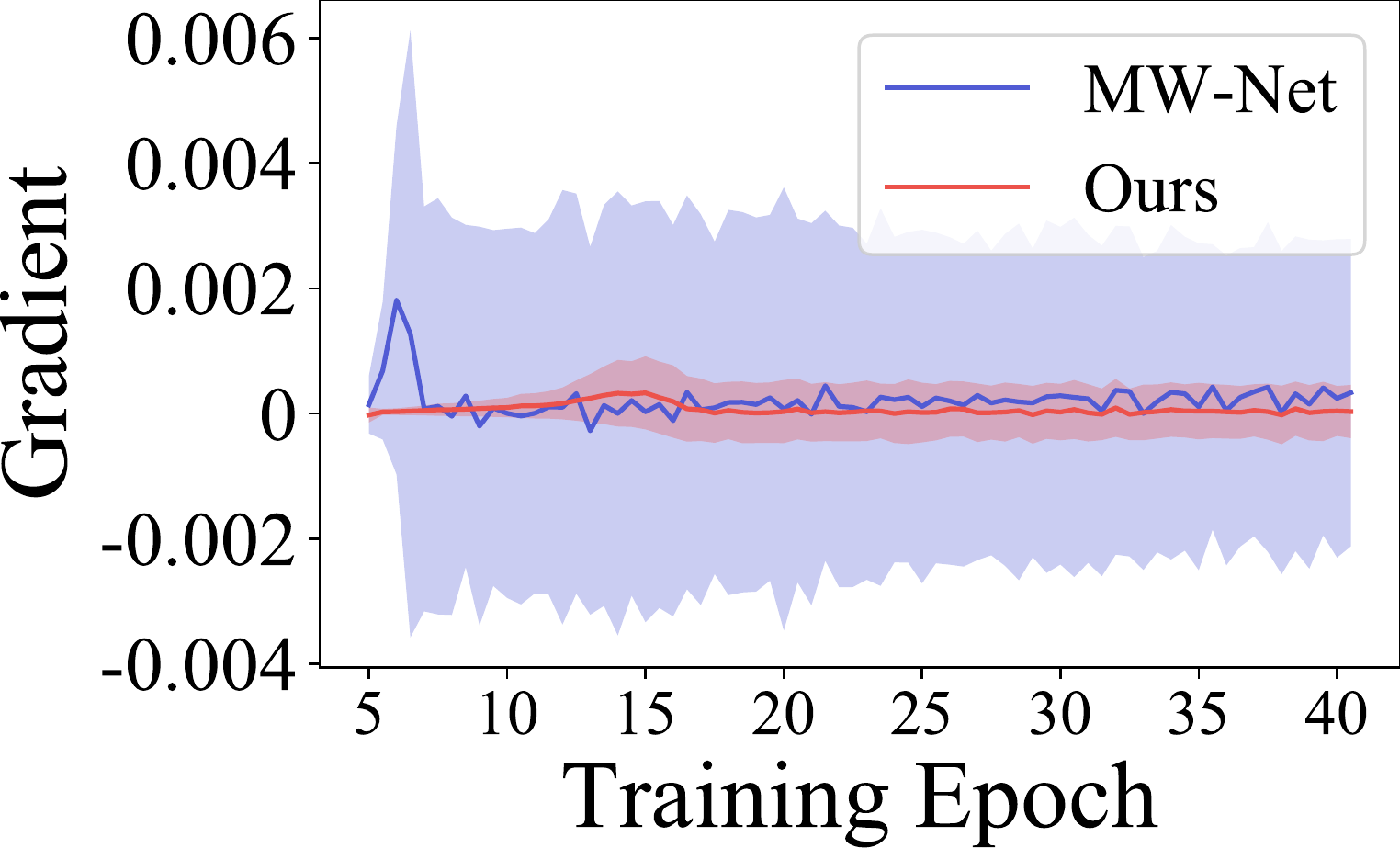}
\footnotesize\caption{Gradient Variance}
\end{subfigure}
\caption{(a) Top-1 accuracy vs. Training time (in hours) on the WebVision dataset~\cite{li2017webvision}. We apply our method on the MW-Net model~\cite{shu2019meta} and train them using the identical hardware platform of four NVIDIA V100 GPUs. (b) The average GPU running time (in seconds) of each step in MW-Net per training iteration. Inception-ResNet V2 is used as the backbone. (c) The meta gradient during the training process. The solid line denotes the mean and the shaded region show the standard deviation.}
\label{fig:motivation}
\end{figure}

\section{Introduction}
Deep neural networks (DNNs) have achieved impressive results in various computer vision applications such as image classification~\cite{krizhevsky2012imagenet,he2016deep}, object detection~\cite{ren2016faster,redmon2016you,liu2016ssd}, and semantic segmentation~\cite{he2017mask}. 
A notable issue is that DNNs are prone to memorizing the training data~\cite{zhang2016understanding, tanaka2018joint}, aggravating training set bias such as noisy training labels~\cite{zhang2016understanding} or imbalanced class distributions~\cite{he2009learning,zhu2020inflated}. This significantly degrades the generalization capabilities and results in skewed classifiers or degenerated feature representations.

Numerous works have been proposed to tackle this issue (\eg~\cite{jiang2017mentornet, han2018co, ren2018learning, li2020dividemix, lin2017focal}). Among them, meta-learning~\cite{ren2018learning, shu2019meta, wang2020training} has recently emerged as an effective framework to mitigate the training data bias. In a nutshell, it employs a meta-model to correct bias by providing a more precise estimation of the training data. The meta-model is updated by stochastic gradient descent using the \emph{meta gradient} (or the high-order gradient) computed on a small proportion of validation data that is assumed available during training\footnote{The extra validation dataset is not a requirement in meta-learning. As in our experiments, we can use a subset of pseudo-labeled training data as the validation data. In this case, no extra labels or data are used.}.
Recently, meta-learning approaches such as L2R~\cite{ren2018learning}, MW-Net~\cite{shu2019meta}, and MLC~\cite{wang2020training} have shown superior performance on several public benchmarks such as CIFAR~\cite{krizhevsky2009learning}, WebVision~\cite{li2017webvision}, and Clothing1M~\cite{xiao2015learning}.

Despite the promising empirical results~\cite{vyas2020learning,shu2020meta}, slow training is currently the bottleneck that prevents meta-learning from being applied in many applications.
The training time of the meta-learning model is approximately 3$\sim$7 times more than the regular DNN training time. For instance, it could take 4 days with 4 NVIDIA V100 GPUs to train MW-Net~\cite{shu2019meta} on a mini subset of WebVision~\cite{li2017webvision,jiang2017mentornet} of only $\sim$50K images.

To understand why the meta-learning approaches are computationally intensive,
we may divide the training into three stages: Virtual-Train, Meta-Train, and Actual-Train~\cite{wang2020training}, where each stage consists of a forward and a backward step. Figure~\ref{fig:motivation}(b) summarizes the GPU time for each stage using a representative meta-learning model called MW-Net~\cite{shu2019meta}. We find more than 80\% of the total computation comes from the Meta-Train backward step in which the \emph{meta gradient} is computed with respect to the loss on the validation data. In this step, the meta gradient is back-propagated through every layer of the network all the way back to the meta-model to update its parameters. Since the regular training does not have such a step, this overhead cost rapidly becomes significant as the number of layers grows in the deep networks.

In this work, we aim at improving the training efficiency of meta-learning while maintaining the generalization capability. We propose a new Meta-Train step, named \methodname\space (\methodnameabbre), to efficiently compute the meta gradient. The plausibility of our method relies on the important finding that the total meta gradient can be reasonably approximated by the meta gradient accumulated from only a few network layers. As a result, instead of accumulating meta gradients from all layers in the Meta-Train step, we design a gradient sampler that is learned to decide, whether or not, to aggregate the meta gradient for each layer. 
When the learnable gradient sampler is turned off, the meta gradient computation is hence circumvented for the corresponding layer. This saves a considerable amount of computation especially when the gradient samplers for lower layers are turned off.

More importantly, we find the meta gradient yielded by the \methodnameabbre\space has lower variance. Figure~\ref{fig:motivation}(c) shows the total meta gradient of the ground-truth (blue curve) and the approximation by the \methodnameabbre\space (red curve). It shows that our approximation is reasonably close to the mean but has a much lower variance. We hypothesize this is because the \methodnameabbre\space learns to select a small number of most informative layers which hence reduces the noisy or redundant signals in the meta gradient. As shown in~\cite{neelakantan2015adding, miller2017reducing}, reduction in gradient variance results in faster and more stable optimization. We observe similar results in our experiments where our method is able to improve the generalization performance of the recent meta-learning methods on noisy training data. 

We conduct extensive experiments to verify the efficiency and efficacy of the proposed method. We demonstrate two benefits of our method in overcoming corrupted training labels. First, it speeds up the recent meta-learning methods~\cite{ren2018learning, shu2019meta, wang2020training} by at least three times while maintaining the comparable or even better generalization performance. For example, Figure~\ref{fig:motivation}(a) shows a faster and better convergence when we applied our method on the MW-Net model. Second, our method achieves new state-of-the-art performance on multiple benchmarks for both synthetic label noise and realistic label noise, including the challenging CNWL benchmark~\cite{jiang2020beyond}.
The comparison is fair as our meta-model is learned without using any extra data. In addition, we also validate our method on the long-tailed recognition task. On the long-tailed CIFAR dataset~\cite{cui2019class}, our method yields competitive performance compared to the recent strong baseline methods.

The contributions of this paper are three-fold. (1) We propose a new \methodname\space to efficiently learn to approximate the meta gradient, which halves two-thirds of the training time of the recent meta-learning methods~\cite{ren2018learning, shu2019meta, wang2020training}.
(2) We empirically show our approach reduces the variance of the meta gradient and improves the generalization performance of the meta-learning model. 
(3) Our method achieves state-of-the-art performance on several benchmarks with noisy labels.

\section{Related Work}

\noindent\textbf{Corrupted/Noisy training labels.} 
Numerous methods have been recently proposed to learn robust deep networks that can overcome corrupted or noisy training labels. These methods address this problem from a variety of directions. For example, several works~\cite{goldberger2016training, patrini2017making, xia2019anchor, tanaka2018joint, yi2019probabilistic, arazo2019unsupervised, wang2020training} modeled the noise distribution or the transition matrix to correct noisy training samples.
Other approaches tried to reduce the weights assigned to noisy samples~\cite{malach2017decoupling, jiang2017mentornet, jiang2020beyond, su2017pose, shu2019meta, yang2019snapshot}. Another effective strategy is to directly identify the clean samples and only select them to train the models~\cite{han2018co, yu2019does, northcutt2019confident, pleiss2020identifying, li2020dividemix, wei2020combating}. Other contributions in this direction include data augmentation~\cite{zhang2017mixup, liang2020simaug, cheng2020advaug}, semi-supervised learning~\cite{hendrycks2018using, vahdat2017toward, li2020dividemix, zhang2020distilling}, \etc. 

Among them, meta-learning~\cite{ren2018learning, shu2019meta, li2019learning, wang2020training} has recently emerged as an effective framework for addressing the noisy labels. These methods all learn a meta-model from clean validation examples but differ in the specific ways to correct the biased training labels. For example, L2R~\cite{ren2018learning} directly adjusts the weight for each example. MLNT~\cite{li2019learning} simulates regular training with synthetic noisy labels. MW-Net~\cite{shu2019meta} learns an explicit weighting function. MLC~\cite{wang2020training} estimates the noise transition matrix.

This paper aims at improving the training efficiency of the meta-learning models. The results show our method not only significantly reduces the training time of three recent meta-learning approaches but also improves their robustness to noisy labels on several standard benchmarks.

\noindent\textbf{Long-tailed recognition.} Long-tailed recognition has been an active research field in computer vision~\cite{chawla2002smote, drummond2003c4, han2005borderline, shen2016relay, mahajan2018exploring, yin2019feature, liu2019large, khan2017cost, cui2019class, cao2019learning, Jamal_2020_CVPR, zhou2020bbn,zhu2020inflated}. 
For example, \cite{chawla2002smote, han2005borderline} aimed to increase the number of minority classes by oversampling, while Drummond~\etal~\cite{drummond2003c4} solved this problem by reducing the number of data in majority classes. Some recent studies~\cite{shen2016relay, mahajan2018exploring} proposed to balance the number of data for each class. \cite{yin2019feature, liu2019large} applied the knowledge learned from the head classes to the tail. 
\cite{khan2017cost, cui2019class, cao2019learning, zhou2020bbn} aimed to manipulate the loss on the class-level based on the data distribution. 

Meta-learning based methods~\cite{ren2018learning, shu2019meta, Jamal_2020_CVPR} have recently achieved promising results on the long-tailed recognition task, in which the meta-model is learned to assign larger weights to the examples of the long-tailed classes. Similar to the noisy labels, meta-learning suffers from slow training speed~\cite{ren2018learning, shu2019meta, wang2020training, Jamal_2020_CVPR}. We show our method improves the efficiency and accuracy of the meta-learning methods on the long-tailed recognition task, and achieves competitive performance compared with recent strong baselines.

\begin{figure*}[t]
\centering
\includegraphics[width=0.94\linewidth]{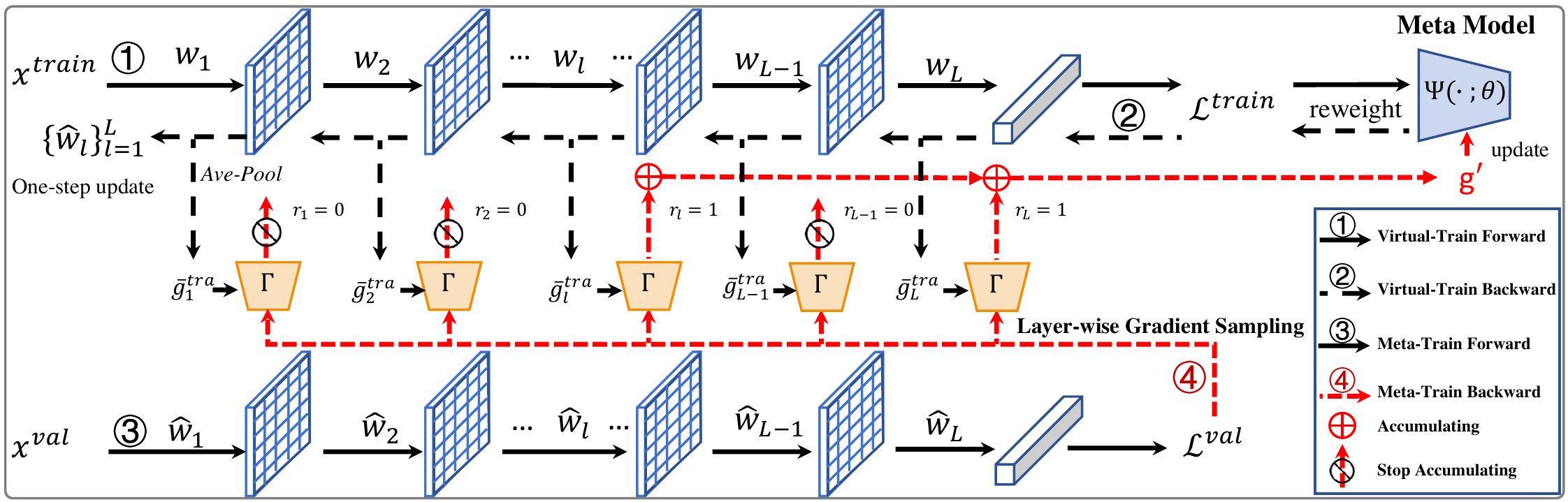}
\caption{Illustration of the proposed method. We propose a new Meta-Train step, named \methodname\space (\ie, the red line \textcircled{4}), which learns a gradient sampler (denoted as $\Gamma$) to aggregate the meta gradient for each layer. In this figure, the meta gradients from the $l$-th and $L$-th layers would be aggregated to compute $\mathbf{g}'$ and used to update the meta-model $\Psi$.}
\vspace{-2mm}
\label{fig:framework}
\end{figure*}

\section{Preliminary on Meta-learning}\label{sec:preliminary}

In this section, we briefly introduce the preliminary on meta-learning methods~\cite{ren2018learning, shu2019meta} that learn robust deep neural networks from noisy labels by reweighting the training data.
We follow the notation in the MW-Net~\cite{shu2019meta} model using corrupted labels as an example. Alternative formulation can be found in~\cite{ren2018learning,wang2020training,vyas2020learning}.

Let $\mathcal{D}^{train}=\{(x_i^{tra}, y_i^{tra})\}_{i=1}^N$ be a noisy training set of $N$ examples, 
where $x_i^{tra}$ is the $i$-th training image and $y_i^{tra}\in \{0, 1\}^{c}$ is its one-hot label over $c$ classes.
Consider a deep neural network (DNN) as the base model $\Phi(\cdot; w)$ with $w$ denoting its parameters.
Generally, we can derive the optimal parameter $w^{*}$ by minimizing the softmax cross-entropy loss $\ell(\hat{y}, y)$ over the training data, where $\hat{y} = \Phi(x; w)$ is the prediction of the DNN and $y$ is the given label for the input image $x$.

In the meta-learning methods~\cite{ren2018learning, shu2019meta}, there is an out-of-sample validation set $\mathcal{D}^{val}=\{(x_j^{val}, y_j^{val})\}_{j=1}^M$, where $(x_j^{val}, y_j^{val})$ denote the $j$-th example. $M$ is the size of $\mathcal{D}^{val}$ and $M \ll N$. The extra validation dataset is not always required in meta-learning. See the discussion in Section~\ref{sec:exp_sota_noisy}. 

The meta-learning method employs a meta-model (\eg, instanced by a multilayer perceptron network (MLP) with only one hidden layer~\cite{shu2019meta}) to learn a weight for each training example. Let $\Psi(\cdot; \theta)$ denote the meta-model, parametrized by $\theta$, which maps a loss to a weight scalar. A meta-model can be regarded as a learnable derivation of the self-paced function in SPCL~\cite{jiang2015self}.
Let $\mathcal{L}_i^{tra}(w) = \ell(\Phi(x_i^{tra}; w), y_i^{tra})$ be the loss for the $i$-th example in $\mathcal{D}^{train}$. The optimal parameter $w^*$ can be obtained by computing the weighted loss: 
\begin{equation}\label{eq:obj_classifer}
    \begin{split}
        w^{*}(\theta) = \argmin_{w} 
        \frac{1}{N}\sum_{i=1}^{N}\mathcal{V}_i^{tra}(\theta)\mathcal{L}_i^{tra}(w),
    \end{split}
\end{equation}
where $\mathcal{V}_i^{tra}(\theta)=\Psi(\mathcal{L}_i^{tra}(w); \theta)$ is the generated weight for the $i$-th training example.

The meta-model is optimized by minimizing the validation loss:
\begin{equation}\label{eq:obj_meta_model}
    \theta^{*} = \argmin_{\theta} 
    \frac{1}{M}\sum_{j=1}^{M}\mathcal{L}_j^{val}(w^{*}(\theta)),
\end{equation}
where $\mathcal{L}_j^{val}(w^{*}(\theta))=\ell(\Phi(x_j^{val}; w^{*}(\theta)), y_j^{val})$ is the loss for the $j$-th example in the validation set.

Solving Eq.~\eqref{eq:obj_classifer} and Eq.~\eqref{eq:obj_meta_model} by alternating minimization is intractable for mini-batch gradient descent. Alternatively, an online optimization method is used instead which comprises three steps: Virtual-Train, Meta-Train, and Actual-Train~\cite{wei2020combating}. 

Consider the $t$-th iteration. Given a training mini-batch $\mathcal{B}^{train}=\{(x_i^{tra}, y_i^{tra})\}_{i=1}^{n}$ and a validation mini-batch $\mathcal{B}^{val}=\{(x_j^{val}, y_j^{val})\}_{j=1}^{m}$, $n$ and $m$ stand for the number of the examples in the mini-batch. 
For the Virtual-Train, an one-step ``virtually'' updated DNN can be derived by:
\begin{equation}\label{eq:virtual_train}
\begin{split}
    \hat{w}(\theta) = w - \alpha \frac{1}{n} \sum_{i=1}^{n}\mathcal{V}_i^{tra}(\theta)\nabla_{w}\mathcal{L}_i^{tra}(w),
\end{split}
\end{equation}
where $\alpha$ is the learning rate for the DNN. $w$ is the parameter of the base DNN at the current iteration. This step is called Virtual-Train because $\hat{w}(\theta)$ will not be used to update the parameter of the base DNN. 

Then for the Meta-Train, with the latest $\hat{w}(\theta)$, the meta-model is updated by:
\begin{equation}\label{eq:meta_train}
\begin{split}
    \theta' = \theta - \beta \frac{1}{m}\sum_{j=1}^{m}\nabla_{\theta}\mathcal{L}_j^{val}(\hat{w}(\theta)).
\end{split}
\end{equation}
Similarly, $\beta$ is the learning rate for the meta-model. $\theta'$ is the parameter of the updated meta-model.
Notice that $\frac{1}{m}\sum_{j=1}^{m}\nabla_{\theta}\mathcal{L}_j^{val}(\hat{w}(\theta))$ is called meta gradient, which is expensive to compute. More details will be discussed in Section~\ref{sec:slow_training}.

Finally, in the last step (Actual-Train), the updated meta-model $\Psi(\cdot;\theta')$ is used to update the base DNN model using:
\begin{equation}\label{eq:actual_train}
\begin{split}
    w' = w - \alpha \frac{1}{n} \sum_{i=1}^{n}\mathcal{V}_i^{tra}(\theta')\nabla_{w}\mathcal{L}_i^{tra}(w),
\end{split}
\end{equation}
where $\mathcal{V}_i^{tra}(\theta')$ is the weight for the $i$-th example computed by the latest meta-model. This step is called Actual-Train because $w'$ will be used to actually update the parameter of base DNN. Therefore, $w'$ becomes the $w$ in Eq.~\eqref{eq:virtual_train} in the $(t+1)$-th iteration.

\section{\methodname}
In this section, we introduce a \methodname\space(\methodnameabbre) to efficiently approximate the total meta gradients by a layer-wise meta gradient sampling procedure. 
Figure~\ref{fig:framework} presents the overall training process, where the red line indicates the proposed method. Specifically, we learn a gradient sampler to decide, whether or not, to aggregate the meta gradient for each layer.
In the following, 
we first explain how the meta gradient can be calculated in a layer-wise fashion in Section~\ref{sec:slow_training}.
Next, we detail the gradient sampler in Section~\ref{sec:LGSM} and the final objective for the meta-model in Section~\ref{sec:overall_objective}. 
The full algorithm for \methodname\space is shown in the supplementary materials.

\subsection{Layer-wise meta gradient computation}\label{sec:slow_training}
In this section, we discuss the meta gradient computation and show how it can be calculated in a layer-wise fashion.
For notational convenience, we simplify Eq.~\eqref{eq:meta_train} as:
\begin{equation}\label{eq:simple_meta_train}
\theta' = \theta - \beta \times \mathbf{g},
\end{equation}
where $\mathbf{g} = \frac{1}{m}\sum_{j=1}^{m}\nabla_{\theta}\mathcal{L}_j^{val}(\hat{w}(\theta))$ denotes the meta gradient, which has shown to be computational intensive in recent studies~\cite{ren2018learning, shu2019meta, wang2020training}.

Without loss of generality, suppose that the base DNN has $L$ layers denoted as $\Phi(\cdot; \{w_l\}_{l=1}^{L})$, where $w_l$ represents the parameter for the $l$-th layer.

We rewrite the computation of meta gradient using the chain rule:
\begin{equation}\label{eq:gradient_accumulate}
\begin{split}
\mathbf{g} &= \frac{1}{m}\sum_{j=1}^{m}\frac{\partial \mathcal{L}_j^{val}(\hat{w}(\theta))}{\partial \hat{w}(\theta)}
  \sum_{i=1}^{n} \frac{\partial \hat{w}(\theta)}{\partial \mathcal{V}_i^{tra}(\theta)}
              \frac{\partial \mathcal{V}_i^{tra}(\theta)}{\partial \theta} \\
           &\propto \frac{-\alpha}{nm}\sum_{l=1}^{L}
                  \bigg(\sum_{i=1}^{n}\bigg(\sum_{j=1}^{m}G_{i,j,l}\bigg)
                  \frac{\partial \mathcal{V}_i^{tra}(\theta)}{\partial \theta}\bigg), 
\end{split}
\end{equation}
where $G_{i,j,l}= (\frac{\partial \mathcal{L}_j^{val}(\hat{w})}{\partial\hat{w}_l})^{\mathsf{T}}\frac{\partial \mathcal{L}_i^{tra}(w)}{\partial w_l}$ is the dot product between 
the gradient from the $j$-th validation loss \wrt $\hat{w}_l$
and the gradient from the $i$-th training loss \wrt $w_l$.
Intuitively, $G_{i,j,l}$ can be viewed as the similarity between the $i$-th training example and the $j$-th validation example according to the $l$-th layer of the base network. 
The derivation of Eq.~\eqref{eq:gradient_accumulate} can be found in the supplementary materials.

Two observations can be drawn from Eq.~\eqref{eq:gradient_accumulate}. First, it explains the slow Meta-Train step in meta-learning, \ie computing the meta gradient involves enumerating all training examples, all validation examples, and all layers. Second, it shows that the meta gradient can be calculated by first computing the gradient $\sum_{i,j} G_{i,j,:}$ within each individual layer and then aggregating the values together. This finding lays a foundation for the proposed layer-wise meta gradient approximation.

\subsection{Layer-wise gradient sampler}\label{sec:LGSM}

We propose to approximate the total meta gradient by aggregating meta gradients sampled from a few layers. We learn a gradient sampler to accumulate the meta gradient for each layer, and formulate the gradient sampler, denoted as $\Gamma(\cdot; \eta_l)$, as follow:
\begin{equation}
\begin{split}
    \label{eq:rl} r_l &= \Gamma(\bar{g}_l^{tra}; \eta_l) \\
    &=\Gamma(\textit{Avg-Pool}(\frac{1}{n}\sum_{i=1}^{n}\mathcal{V}_i^{tra}(\theta)\frac{\partial \mathcal{L}_i^{tra}(w)}{\partial w_l}); \eta_l).
\end{split}
\end{equation}
The output to the gradient sampler is the discrete activation status $r_l \in \{0, 1\}$.

The input of the gradient sampler is the average gradient $\bar{g}_l^{tra}$ obtained from the Virtual-Train backward step.
To be more specific, suppose the gradient tensor for the convolutional kernel has the shape $\mathbb{R}^{D_{out}\times D_{in} \times K_1 \times K_2}$ where $D_{out}$ and $D_{in}$ are the output/input dimensions; $K_1$ and $K_2$ are the kernel sizes. The $\textit{Avg-Pool}$ operator averages the gradient tensor across all except the first dimensions while leaving the bias term unchanged. Therefore, the dimension for $\bar{g}_l^{tra} \in \mathbb{R}^{1\times D_{out}}$.
The $\textit{Avg-Pool}$ performs a similar operation for the fully connected layer by setting $K_1=K_2=1$.

For efficiency, we adopt a lightweight design for the gradient sampler and implement it by two fully-connected (FC) layers: FC$_1$ and FC$_2$, where the first layer FC$_1$ is followed by a PReLU layer and FC$_2$ by the ``Gumbel-softmax'' operator~\cite{jang2016categorical}. The hidden size of the fully connected layer is fixed to 128 for all experiments.

Applying the gradient sampler to all layers gives:
{\small
\begin{align}\label{eq:our_meta_gradient}
    &\mathbf{g}' \propto \frac{-\alpha}{nm}\sum_{l=1}^{L} \mathbbm{1}_{[r_l=1]}
                      \bigg(\sum_{i=1}^{n}\bigg(\sum_{j=1}^{m}G_{i,j,l}\bigg)
                      \frac{\partial \mathcal{V}_i^{tra}(\theta)}{\partial \theta}\bigg), 
\end{align}
}where $\mathbbm{1}_{[r_l=1]}$ is the indicator function.
As shown in Eq.~\eqref{eq:our_meta_gradient}, the meta gradient for the $l$-th layer is accumulated only if the gradient sampler is turned on (\ie $r_l=1$).

Finally, we replace $\mathbf{g}$ in Eq.~\eqref{eq:simple_meta_train} with  $\mathbf{g}'$ to update the meta-model.

\subsection{Training objective for meta-model}\label{sec:overall_objective}
The proposed gradient samplers are jointly optimized with the meta-model. In addition to the cross-entropy loss described in $\mathcal{L}^{val}$ in Eq.~\eqref{eq:meta_train}, we incorporate two auxiliary losses to facilitate learning the gradient samplers.

The first loss is designed to prevent the gradient samplers from activating too many layers. We introduce a loss $\mathcal{L}_{r}$ regularizing the output of the gradient samplers:
\begin{align}
\label{eq:l_r}
    \mathcal{L}_{r} &= \|\sum_{l=1}^L {r_l} - K\|_2^{2},
\end{align}
where $K$ is the expected number of layers to be activated. 

Moreover, we add another loss (denoted as $\mathcal{L}_{g}$) to facilitate learning the meta-model:
\begin{align}
    \mathcal{L}_{g} &= \|\bar{g}_{L}^{tra} - \bar{g}_{L}^{val}\|_2^{2},
\end{align}
where $\bar{g}_{L}^{tra}$ is the average gradient from training loss discussed in Eq.~\eqref{eq:rl}. Likewise $\bar{g}_{L}^{val}$ is the average gradient from the validation loss, \ie $\bar{g}_{L}^{val}= \textit{Avg-Pool}(\frac{1}{m}\sum_{j=1}^{m}\frac{\partial \mathcal{L}_j^{val}(\hat{w})}{\partial\hat{w}_L})$. This loss term $\mathcal{L}_{g}$ captures the prior knowledge that the distance between validation and training gradient should be close. Notice that we only compute the gradients at the last layer $L$ for efficiency.

Finally, the total loss to update the meta-model:
\begin{align}
\label{eq:final_loss}
    \mathcal{L}^{val} &= \mathcal{L}_{c} + \lambda_1 \mathcal{L}_{r} + \lambda_2 \mathcal{L}_{g},
\end{align}
where $\mathcal{L}_{c}$ is the standard cross-entropy loss in Eq.~\eqref{eq:meta_train}. 
$\lambda_1$ and $\lambda_2$ are hyperparameters. We will examine the effectiveness of these loss terms in the ablation study.

\section{Experiments}
We conduct extensive experiments on the noisy labeled data to verify the efficiency and effectiveness of our method for learning robust DNN models. Specifically, we show our method improves the efficiency and generalization performance of the meta-learning methods in Section~\ref{sec:comp_meta_methods}. Section~\ref{sec:ablation_study} presents ablation studies to verify our design choices.
Section~\ref{sec:exp_sota_noisy} compares with the state-of-the-art results on synthetic and realistic noisy labels. In addition, we also experiment on the long-tailed recognition task in Section~\ref{sec:exp_longtail}.
The implementation details and more experimental results are presented in the supplementary materials.

\begin{table}[t]
	\centering
	\footnotesize
	\begin{tabular}{|l|p{1.0cm}|p{0.4cm}p{0.4cm}p{0.4cm}|p{0.4cm}p{0.4cm}p{0.4cm}|}
	\hline
	\multirow{2}{*}{Method}  & \multirow{2}{*}{Time (ms)} & \multicolumn{3}{|c|}{CIFAR-10} & \multicolumn{3}{|c|}{CIFAR-100} \\
	\cline{3-8}
	& & 20\% & 40\% & 60\% & 20\% & 40\% & 60\% \\
	\hline
	\small{MW-Net~\cite{shu2019meta}} & 933 & 91.9 & 89.6 & 84.5 & 73.1 & 68.1 & 61.7  \\
	\hline
	\textbf{+FaMUS} & 284(\textcolor{red}{3.3x}) & 92.9 & 90.5 & 85.8 & 73.6 & 69.4 & 62.9  \\
	\hline\hline
	L2R~\cite{ren2018learning}& 839 & 90.5 & 86.9 & 82.2 & 69.3 & 62.8 & 50.8  \\
	\hline
	\textbf{+FaMUS} & 244(\textcolor{red}{3.4x}) & 91.3 & 87.6 & 82.8 & 70.7 &	65.5 & 51.6  \\
	\hline
	\end{tabular}
	\vspace{-2mm}
	\caption{Comparison with MW-Net and L2R on CIFAR-10 and CIFAR-100. Percentage numbers represent the noise rate. ``Time (ms)'' denotes the average running time per training iteration on a single NVIDIA V100 GPU. 
	}\label{tab:comp_mwn_and_l2r}
\end{table}

\iftrue
\begin{table}[t]
	\centering
	\small
	\begin{tabular}{|l|l|cccc|}
		\hline
		Method & Time (ms) & 10\% & 20\% & 30\% & 40\% \\
		\hline
		MLC~\cite{wang2020training} & 265 & 85.23 & 84.28 & 82.10 & 79.89 \\ 
		\hline
		\textbf{+FaMUS} & 84(\textcolor{red}{3.1x}) & 87.28 & 85.00 & 82.65 & 80.41 \\
		\hline
	\end{tabular}
	\vspace{-2mm}
	\caption{Comparison with MLC on CIFAR-10 with four different noise rates: \{10\%, 20\%, 30\%, 40\%\}. ``Time (ms)'' denotes the average running time per training iteration on a single NVIDIA V100 GPU.}\label{tab:cifar10_mlc_uniform}
	\vspace{-5mm}
\end{table}
\fi

\subsection{Comparison with meta-learning methods}\label{sec:comp_meta_methods}

This subsection shows our method improves the efficiency and generalization performance of three meta-learning methods: L2R~\cite{ren2018learning}, MW-Net~\cite{shu2019meta}, and MLC~\cite{wang2020training}. 

\textbf{Setups.} We apply our method to three meta-learning methods using their official code and train them under the same settings as reported in their papers~\cite{ren2018learning, shu2019meta, wang2020training}. This includes using the same clean validation set to learn the meta-model. The experiments are conducted on the standard CIFAR~\cite{krizhevsky2009learning} benchmarks. Following~\cite{wang2020training}, we use the \textit{symmetric} label noise in which a percentage of true labels are randomly replaced with all possible labels, and report the best peak accuracy which is the maximum accuracy on the clean test set during training. 

\textbf{Implementation details.} The proposed gradient samplers are jointly optimized with the meta-model by SGD with a momentum of 0.9. The learning rate is fixed as 0.1 throughout the training. 
$\lambda_1$ and $\lambda_2$ are both set to 0.1. $K$ is set to 4.

\begin{figure}[t]
\begin{subfigure}{.23\textwidth}
\includegraphics[width=1.0\linewidth]{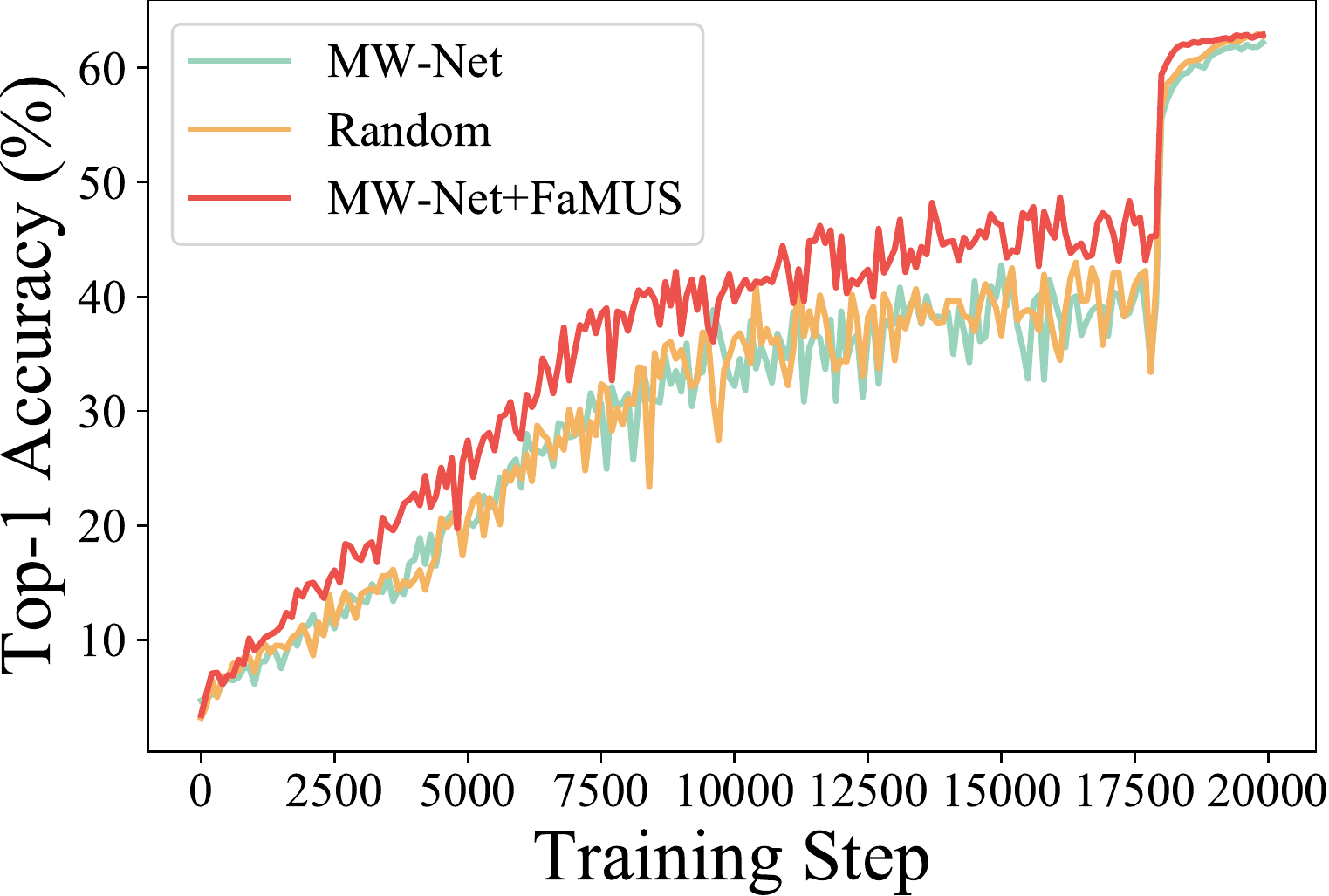}
\footnotesize\caption{Top-1 Acc vs. Training Step}
\end{subfigure}
\begin{subfigure}{.23\textwidth}
\includegraphics[width=1.0\linewidth]{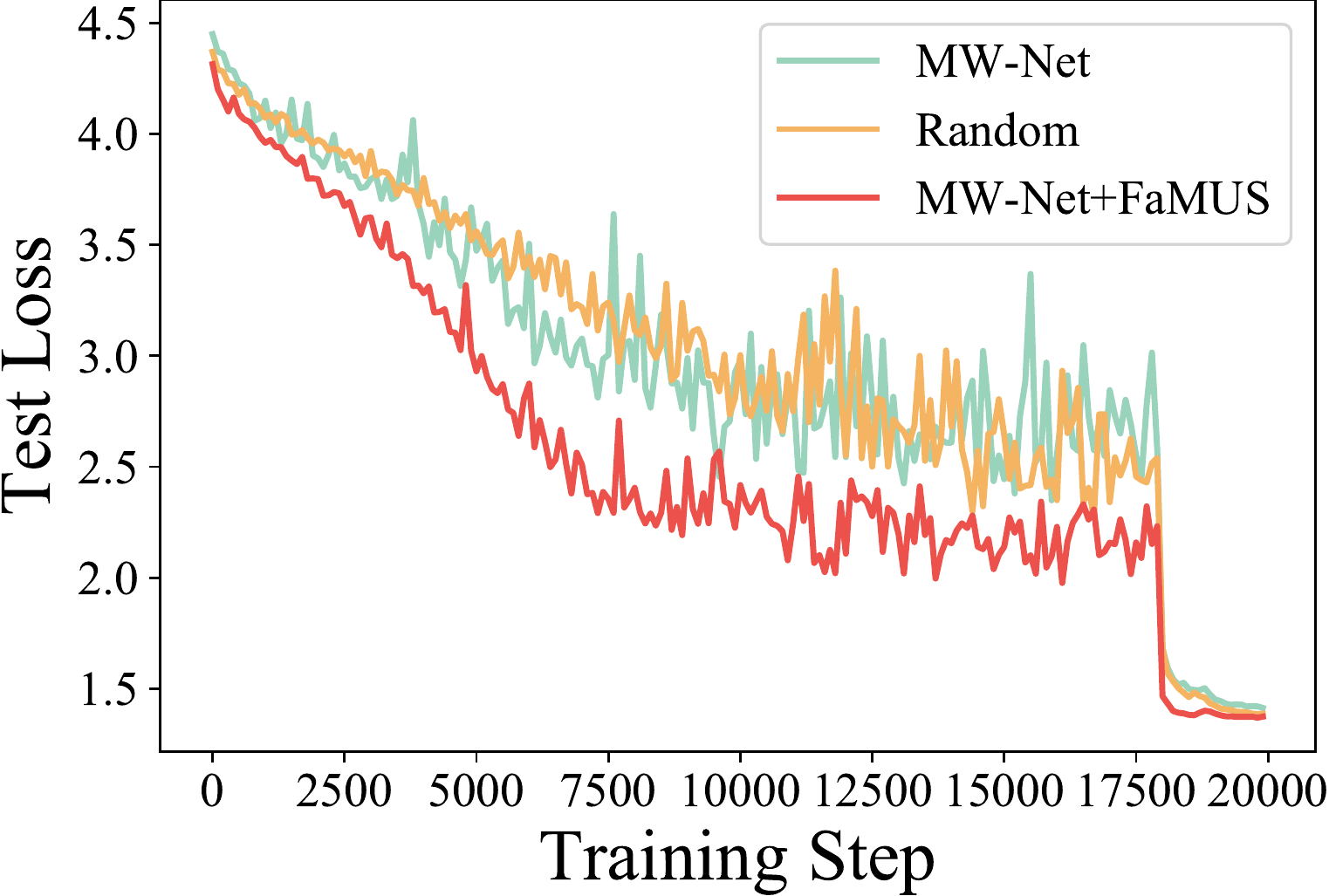}
\footnotesize\caption{Test loss vs. Training Step}
\end{subfigure}
\vspace{-2mm}
\caption{Test curves under CIFAR-100 with 60\% noise.}
\label{fig:test_acc}
\end{figure}

Table~\ref{tab:comp_mwn_and_l2r} and Table~\ref{tab:cifar10_mlc_uniform} show the results on the CIFAR datasets, where ``Time'' column lists the average running time (in millisecond) per training iteration on a single NVIDIA V100 GPU. It shows that our method accelerates the training time of the three meta-learning methods~\cite{ren2018learning, shu2019meta, wang2020training} by at least 3 times. More importantly, our method improves their generalization performance across all noise rates. To understand the training dynamics, we compare three methods: the best baseline MW-Net~\cite{shu2019meta}, the MW-Net with our method, and the Random MW-Net in which each layer is randomly sampled to compute the meta gradient.
Figure~\ref{fig:test_acc} shows the training curves on the CIFAR-100 dataset with 60\% noise. We observe that our method (MW-Net + FaMUS) has the lowest test loss and the highest accuracy throughout the training.

\begin{figure}[t]
\centering
\includegraphics[width=0.95\linewidth]{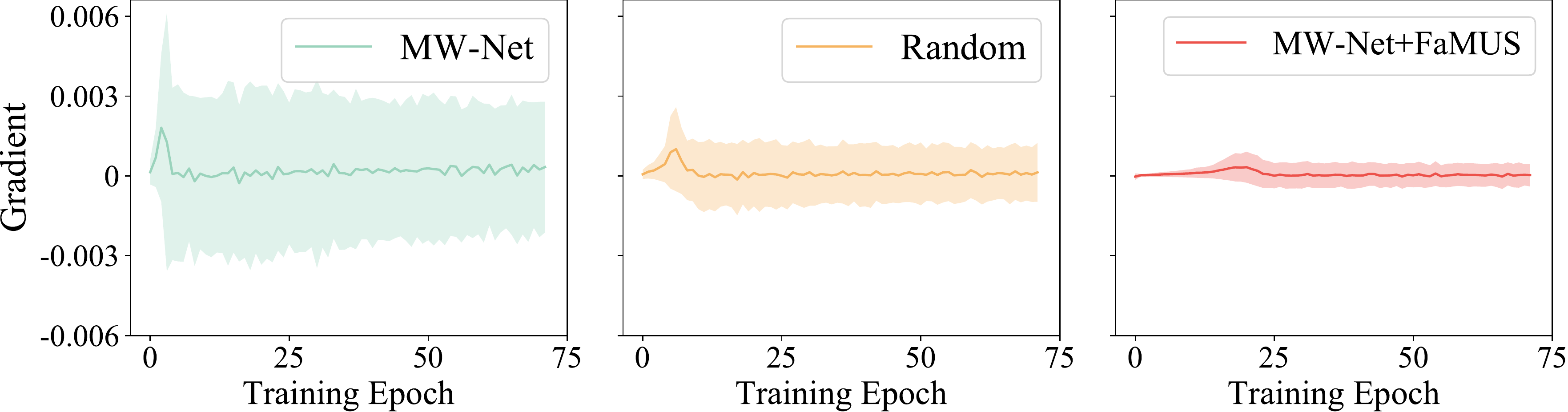}
\vspace{-2mm}
\caption{Variance of the meta gradient produced by different methods during the training. All models are trained on the CIFAR-100 dataset with 60\% noise.}
\label{fig:weight_variation}
\end{figure}

\begin{figure}[t]
\centering
\begin{subfigure}{.23\textwidth}
\includegraphics[width=1.0\linewidth]{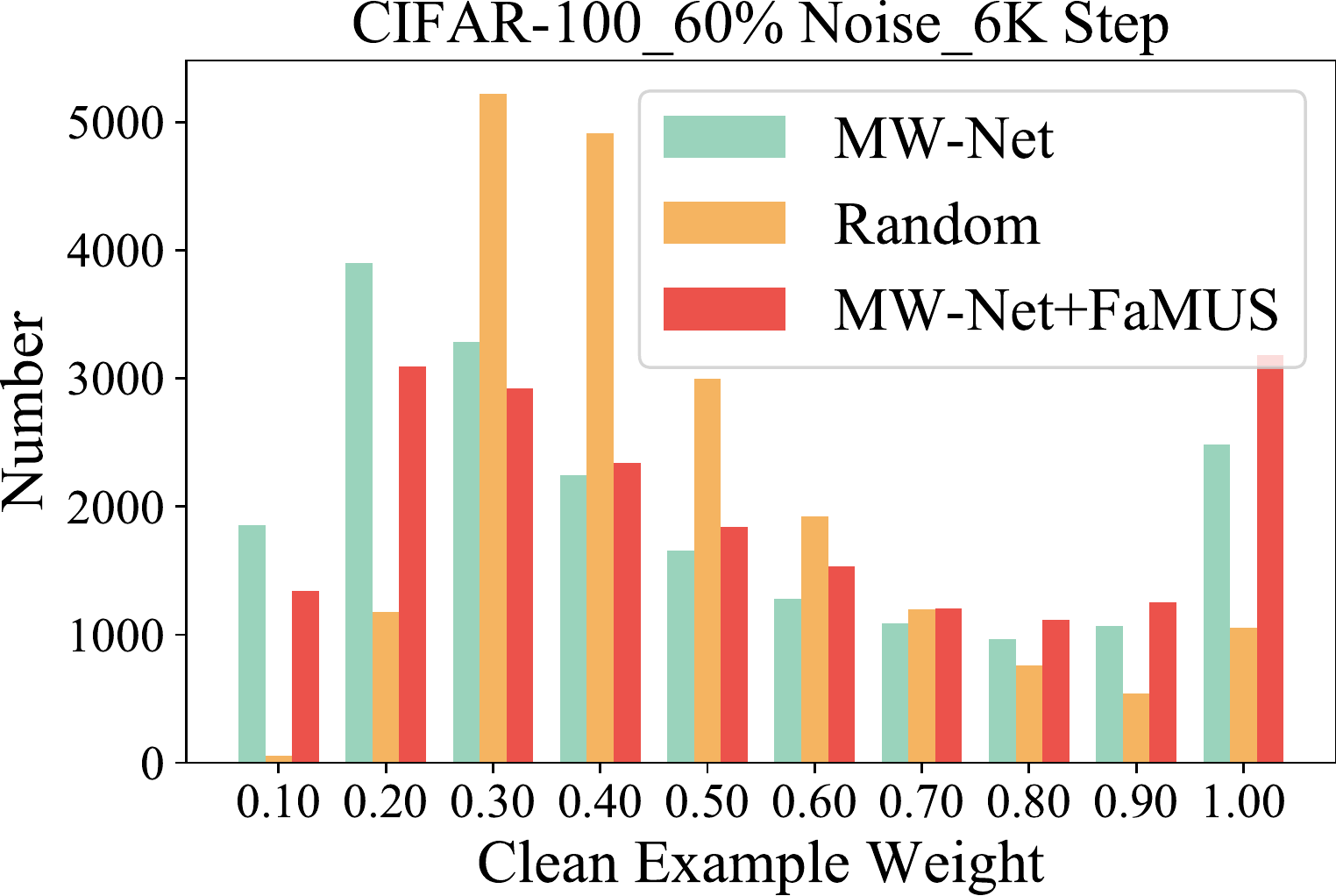}
\end{subfigure}
\begin{subfigure}{.23\textwidth}
\includegraphics[width=1.0\linewidth]{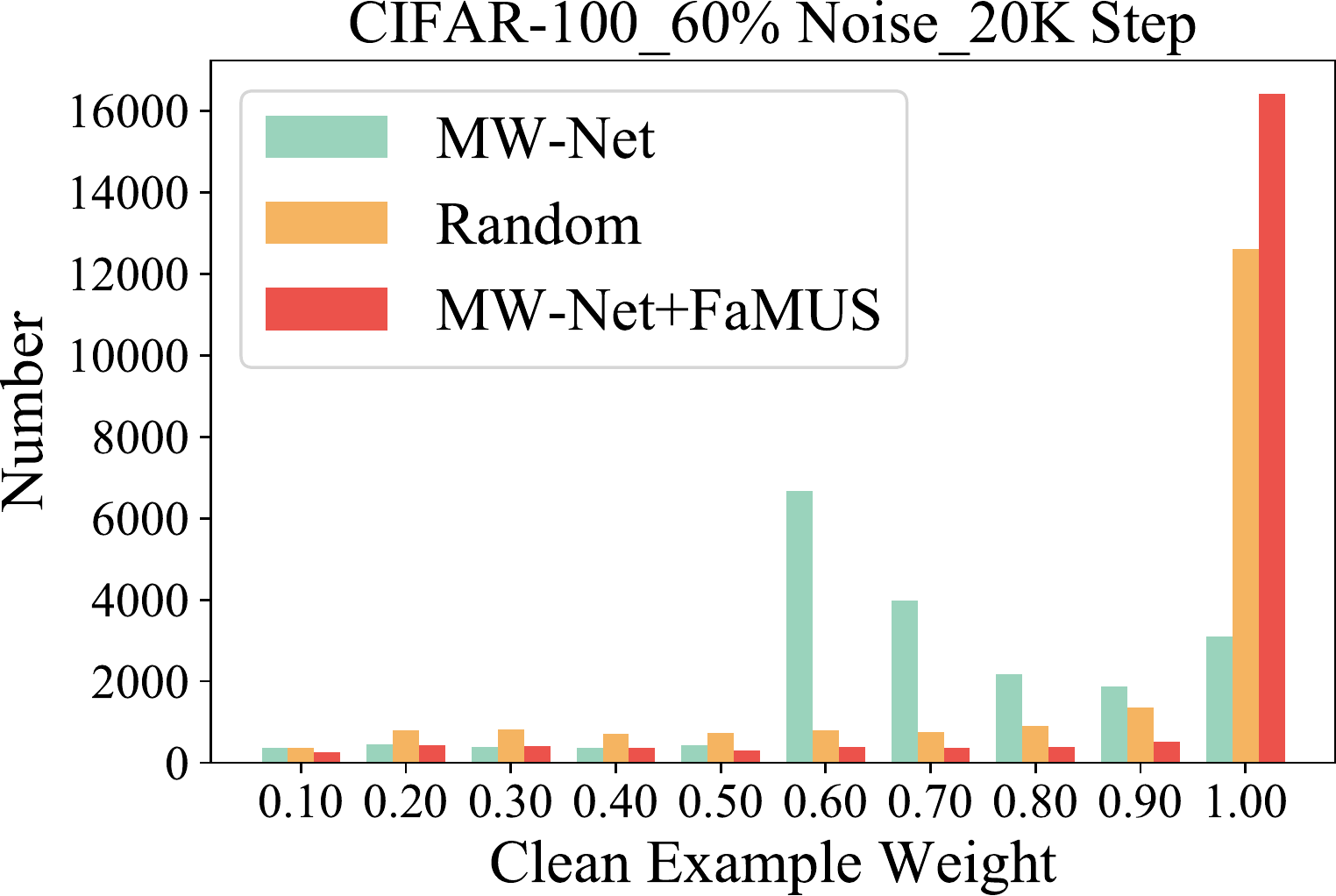}
\end{subfigure}
\vspace{-2mm}
\caption{Weight distribution over the clean examples in the \textit{6K} (left) and \textit{20K} (right) training step. All models are trained on the CIFAR-100 dataset with 60\% noise.}
\label{fig:weight_distutions}
\vspace{-6mm}
\end{figure}

We hypothesize that this benefit is related to the low variance in the meta gradient learned by our method. For example, Figure~\ref{fig:weight_variation} visualizes the variance of the meta gradient during training and shows our method yields a low-variance approximation of the meta gradient. These results suggest that \methodnameabbre\space can learn to select a small number of most informative layers to compute the meta gradient, which hence reduces the noisy learning signals in the corrupted training data. This observation agrees with the finding in~\cite{neelakantan2015adding, miller2017reducing} that reduction in gradient variance results in faster and more stable optimization.
To substantiate this hypothesis, we examine the meta-models by plotting their weight distribution on the clean examples in Figure~\ref{fig:weight_distutions}. We find that in both the early (\textit{6K} step) and late training stages (\textit{20K} step), the meta-models learned by our method tend to assign larger weights to more clean examples.
The results in Table~\ref{tab:comp_mwn_and_l2r}, Table~\ref{tab:cifar10_mlc_uniform}, and  Figure~\ref{fig:test_acc} demonstrate that our method improves the efficiency and generalization performance of the meta-learning methods. More results can be found in the Appendix C.

\begin{table}[t]
	\centering
	\small
	\begin{tabular}{|ccc|c|c|c|}
		\hline
		$\mathcal{L}_c$ & $\mathcal{L}_r$ & $\mathcal{L}_g$ & Time (ms) &CIFAR-10 & CIFAR-100 \\
		\hline
		\checkmark &  &                      & 933  & 84.5  & 61.7 \\
		\hline
		\checkmark &          & \checkmark   & 446  & 85.2 & 62.2 \\
		\hline
		\checkmark & \checkmark &            & 275  & 85.3 & 62.0   \\
		\hline
		\checkmark & \checkmark & \checkmark & 284   & 85.8 & 62.9 \\
		\hline
	\end{tabular}
	\vspace{-1mm}
	\caption{Accuracy vs. Training Time on CIFAR-10 and CIFAR-100 with 60\% noise. 
	``Time (ms)'' denotes the average running time per training iteration on a single NVIDIA V100 GPU. 
	}\label{tab:ablation_loss_function}
\end{table}

\begin{table}[t]
	\centering
	\small
	\begin{tabular}{|l|l|c|c|}
		\hline
		\multicolumn{2}{|c|}{Model} & Time (ms) & ACC   \\
		\hline
        \multicolumn{2}{|c|}{MW-Net~\cite{shu2019meta}} & 933 & 61.7  \\
		\hline
		\multirow{3}{*}{ Pre-specified Block} & $b$ = 4  & 718 & 60.8  \\
                                           & $b$ = 8  & 569 & 61.9  \\
                                           & $b$ = 12 & 326 & 61.4  \\
        \hline
        \multirow{3}{*}{ Random Layers} & $s$ = 4  & 764 & 61.2  \\
                                       & $s$ = 8  & 826 & 61.6  \\
                                       & $s$ = 16 & 899 & 62.1  \\
		\hline\hline
		\multicolumn{2}{|c|}{\textbf{FaMUS} ($K=4$)} & \textbf{284} & \textbf{62.9}  \\  
		\hline
	\end{tabular}
	\vspace{-1mm}
	\caption{Comparison of sampling strategies on CIFAR-100 with 60\% noise. $b\in[1, 12]$ is the index of the residual block. $s$ is the number of randomly selected layers. Our method samples from about 4 layers by setting $K=4$ in Eq.~\eqref{eq:l_r}. ``Time (ms)'' denotes the average running time per training iteration on a single NVIDIA V100 GPU.
	}\label{tab:ablation_diff_accumulation}
	\vspace{-5mm}
\end{table}

\subsection{Ablation study}\label{sec:ablation_study}
We conduct the ablation studies using the MW-Net method with the WideResNet-28-10 backbone model~\cite{zagoruyko2016wide}.

\textbf{Loss function.} Table~\ref{tab:ablation_loss_function} analyzes the impact of the auxiliary loss components in Eq.~\eqref{eq:final_loss} on the CIFAR datasets with 60\% noise. ``$\mathcal{L}_c$'' denotes the loss of the base meta-learning model (MW-Net).
We find that adding ``$\mathcal{L}_r$'' significantly reduces the training time since it limits the number of layers to be activated. 
When incorporating both ``$\mathcal{L}_r$'' and ``$\mathcal{L}_g$'', our method achieves the best result and improves both the efficiency and accuracy of the base MW-Net model.

\textbf{Sampling strategy.} To verify the design of the proposed gradient sampler, we compare with two predefined sampling strategies: \textit{pre-specified block} and \textit{random layers}. In the Pre-specified Block, we select a residual block (indexed by $b$ and $b\in[1, 12]$), which consists of two convolutional layers and two batch normalization layers, to compute the meta gradients. In the Random Layers, we uniformly select $s$ layers to compute the meta gradients.

Table~\ref{tab:ablation_diff_accumulation} shows the comparison on the CIFAR-100 dataset with 60\% noise rate. For the Pre-specified Block, we find that computing the meta gradient from the top residual block ($b=12$) is the most efficient way, which is about $2$x faster than using the bottom block ($b=4$). 
As for the Random Layers, the accuracy is improved as the number of layers $s$ increases, while the running time shows a different trend.
Our method outperforms all the compared methods both in efficiency and accuracy, suggesting the necessity of the proposed gradient sampler. 

\begin{table}[t]
	\centering
	\small
	\begin{tabular}{|lr|cc|cc|}
		\hline
		\multicolumn{2}{|c|}{\multirow{2}{*}{Method}}  & \multicolumn{2}{|c|}{CIFAR-10} & \multicolumn{2}{|c|}{CIFAR-100} \\
		\cline{3-6}
		      &  & 40\% & 60\% & 40\% & 60\% \\
		\hline

        \multicolumn{2}{|l|}{Co-teaching~\cite{han2018co}} & 74.81 & 73.06 & 46.20 & 35.67 \\  
        \hline
        \multicolumn{2}{|l|}{L2R~\cite{ren2018learning}}& 86.92 & 82.24 & 62.81 & 50.81 \\
        \hline
        \multicolumn{2}{|l|}{MW-Net~\cite{shu2019meta}} & 89.60 & 84.49 & 68.11 & 61.71 \\
        \hline
        \multicolumn{2}{|l|}{MentorNet$^{\dagger}$~\cite{jiang2017mentornet}} & 91.20 & 74.20 & 66.80 & 58.80 \\
        \hline
        \multicolumn{2}{|l|}{Mixup$^{\dagger}$~\cite{zhang2017mixup}} & 91.50 & 86.80 & 66.80 & 58.80 \\
        \hline
        \multicolumn{2}{|l|}{M-correction~\cite{arazo2019unsupervised}} & 92.80 & 90.30 & 70.10 & 59.50 \\
        \hline
        \multicolumn{2}{|l|}{MentorMix~\cite{jiang2020beyond}} & 94.20 & 91.30 & 71.30 & 64.60 \\
        \hline
        \multicolumn{2}{|l|}{DivideMix~\cite{li2020dividemix}} & 94.90 & 94.30 & 75.20 & 72.00 \\
		\hline\hline
        \multicolumn{2}{|l|}{\textbf{Ours}} &  \textbf{95.37} & \textbf{94.97} & \textbf{75.91} & \textbf{73.58}  \\
                   &   &  {\footnotesize $\pm$0.15} & {\footnotesize $\pm$0.11} & {\footnotesize $\pm$0.19} & {\footnotesize $\pm$0.28}  \\
		\hline
	\end{tabular}
	\vspace{-1mm}
	\caption{Comparison with the state-of-the-art on CIFAR-10 and CIFAR-100 with 40\% and 60\% noise rates. $^{\dagger}$ denotes the results are reported by~\cite{jiang2020beyond}.}
	\label{tab:cifar_mwn}
	\vspace{-5mm}
\end{table}

\subsection{Comparison to state-of-the-art}\label{sec:exp_sota_noisy}

This subsection compares our method with the state-of-the-art robust learning methods in overcoming both \textit{synthetic} and \textit{realistic noisy labels}.

\textbf{Datasets.} For the realistic noisy labels, we employ three datasets: (mini) WebVision 1.0~\cite{li2017webvision},  Clothing1M~\cite{xiao2015learning}, and \textit{Controlled Noisy Web Labels} (CNWL)~\cite{jiang2020beyond}. \textbf{WebVision} contains 2.4 million images with noisy labels categorized into the same 1,000 classes as in the ImageNet ILSVRC12. Following the previous works~\cite{jiang2017mentornet, chen2019understanding}, we use the first 50 classes of the Google image subset as the training data. 
\textbf{Clothing1M} has 1 million noisy labeled clothing images crawled from online shopping websites. \textbf{CNWL} is a recent benchmark of controlled label noise from the web. Uniquely, it allows for comparing methods on various rates of realistic label noises. We use the Red Mini-ImageNet set~\cite{vinyals2016matching} that consists of 50K images from 100 classes for training and 5K images for testing. 

\textbf{Implementation details.} To fairly compare with the the-state-of-art, we use a subset of pseudo labeled training data as the meta-learning validation set. Inspired by~\cite{li2020dividemix}, we employ the Gaussian Mixture Model (GMM) to divide the training data into a pseudo-clean and a pseudo-noisy label set. 
By doing so, no extra clean labels nor data are used to train the meta-model. We find using the pseudo validation set notably improves the performance because the pseudo validation set is much larger than the clean validation set used in the meta-learning method~\cite{shu2019meta}. 
More discussions can be found in the Appendix E.

For experiments on CIFAR-10, CIFAR-100, and CNWL, we employ the PreAct ResNet-18~\cite{he2016identity} as the base DNN. For experiments on Clothing1M and WebVision, we use ResNet-50~\cite{he2016deep} and Inception-ResNet V2~\cite{szegedy2016inception}, respectively. 

\begin{table}[t]
	\centering
	\small
	\begin{tabular}{|l|c|c|c|c|}
		\hline
		\multirow{2}{*}{Method} & \multicolumn{2}{|c|}{WebVision} & \multicolumn{2}{|c|}{ILSVRC12}\\
		                        \cline{2-5}
		                        & top1 & top5 & top1 & top5 \\
        \hline		                        
	    F-correction~\cite{patrini2017making} & 61.12 & 82.68 & 57.39 & 82.36 \\
        \hline
        Decoupling~\cite{malach2017decoupling} & 62.54 & 84.74 & 58.26 & 82.26 \\
        \hline
        D2L~\cite{ma2018dimensionality} & 62.68 & 84.00 & 57.80 & 81.36 \\
        \hline
        MentorNet~\cite{jiang2017mentornet} & 63.00 & 81.40 & 57.80 & 79.92 \\
        \hline
        Co-teaching~\cite{han2018co} & 63.58 & 85.20  & 61.48 & 84.70 \\
        \hline
        Iterative-CV~\cite{chen2019understanding} & 65.24 & 85.34 & 61.60 & 84.98 \\
        \hline
		MW-Net~\cite{shu2019meta} & 74.52 & 88.89 & 72.60 & 88.80 \\
		\hline
        MentorMix~\cite{jiang2020beyond} & 76.00 & 90.20 & 72.90 & 91.10 \\ 
		\hline
        DivideMix~\cite{li2020dividemix} & 77.32 & 91.64 &	75.20 & 90.84 \\

		\hline\hline
        \textbf{Ours} & \textbf{79.40} & \textbf{92.80} & \textbf{77.00} & \textbf{92.76} \\
		\hline
	\end{tabular}
	\vspace{-1mm}
	\caption{Comparison with the state-of-the-art on (mini) WebVision dataset. Numbers denote top-1 (top-5) accuracy on the validation set of WebVision and ImageNet ILSVRC12.
	}\label{tab:webvision}
\end{table}

\begin{table}[t]
	\centering
	\small
	\begin{tabular}{|l|cccc|c|}
		\hline
		Method & 20\% & 40\% & 60\% & 80\% & Mean\\
		\hline
		Cross-entropy & 47.36 & 42.70 & 37.30 & 29.76 & 39.28 \\
		\hline
		Mixup~\cite{zhang2017mixup} & 49.10 & 46.40 & 40.58 & 33.58 & 42.41 \\
		\hline
		DivideMix~\cite{li2020dividemix} & 50.96 & 46.72 & 43.14 & 34.50 & 43.83 \\ 
		\hline
		MentorMix~\cite{jiang2020beyond} & 51.02 & 47.14 & 43.80 & 33.46 & 43.85 \\
		\hline\hline
		\textbf{Ours (FaMUS)} & \textbf{51.42} & \textbf{48.06} & \textbf{45.10} & \textbf{35.50} & \textbf{45.02} \\
		\hline
	\end{tabular}
	\vspace{-1mm}
	\caption{Results on Controlled Noisy Web Labels~\cite{jiang2020beyond}. 
	}\label{tab:cnwl}
	\vspace{-6mm}
\end{table}

\begin{table*}[ht]
	\centering
	\small
	\begin{tabular}{|l|c|c|c|c|c|c|c|c|}
		\hline
		\multirow{2}{*}{Method} & \multicolumn{4}{c|}{Long-Tailed CIFAR-10} & \multicolumn{4}{c|}{Long-Tailed CIFAR-100} \\
		\cline{2-9}
		 & 100 & 50 & 20 & 10 & 100 & 50 & 20 & 10 \\
		\hline
		CE loss & 70.36 & 74.81 & 82.23 & 86.39 & 38.32 & 43.85 & 51.14 & 55.71 \\
		\hline
		Focal Loss$^{\dagger}$~\cite{lin2017focal} & 70.38 & 76.71 & 82.76 & 86.66 & 38.41 & 44.32 & 51.95 & 55.78 \\
		\hline
		CB Focal$^{\dagger}$~\cite{cui2019class} & 74.57 & 79.27 & 84.36 & 87.49 & 39.60 & 45.32 & 52.59 & 57.99 \\
		
		\hline
		LDAM-DRW~\cite{cao2019learning}  & 77.03 & - & - & 88.16 & 44.70 & - & - & \underline{59.59} \\
		\hline
		BBN~\cite{zhou2020bbn} & 79.82 & 82.18 & - & 88.32 & 42.56 & 47.02 & - & 59.12 \\
		\hline\hline
		
		L2R$^{\dagger}$~\cite{ren2018learning} with CE loss & 74.16 & 78.93 & 82.12 & 85.19 & 40.23 & 44.44 & 51.64 & 53.73 \\
 		\hline
		MW-Net~\cite{shu2019meta} with CE loss & 75.21 & 80.06 & 84.94 & 87.84 & 42.09 & 46.74 & 54.37 & 58.46 \\ \hline                       		                        
		\cite{Jamal_2020_CVPR} with CE loss  &    76.41 & 80.51 & \underline{86.46} & \underline{88.85} &  43.35 & 48.53 & 55.62 & 59.58 \\
        \hline
		\cite{Jamal_2020_CVPR} with LDAM  & \underline{80.00} & 82.34 & 84.37 & 87.40 & 44.08 & 49.16 & 52.38 & 58.00 \\
		\hline
        \textbf{MW-Net with CE loss + FaMUS} & 79.30 & \underline{83.15} & \textbf{87.15} & \textbf{89.39} & \underline{45.60}	 & \underline{49.56} & \textbf{56.22} & \textbf{60.42} \\
		\hline
		\textbf{MW-Net with LDAM loss + FaMUS} & \textbf{80.96} & \textbf{83.32} & 86.24 & 87.90 & \textbf{46.03} & \textbf{49.93} & \underline{55.95}  & 59.03 \\
		\hline
	\end{tabular}\vspace{-3mm}
	\caption{Top-1 test accuracy of ResNet-32 on the long-tailed CIFAR-10 and CIFAR-100 with four imbalanced factors $\{100, 50, 20, 10\}$. Methods in the bottom block use extra clean data. The best performance is in \textbf{bold} and the second best is \underline{underscored}. $^{\dagger}$ denotes the results are reported by~\cite{cao2019learning}. }\label{tab:long_tailed_cifar}
	\vspace{-2mm}
\end{table*}

\textbf{Baselines.} We briefly introduce the baselines:
(1) \textbf{Co-teaching}~\cite{han2018co}, \textbf{Decoupling}~\cite{malach2017decoupling}, and \textbf{JoCoR}~\cite{wei2020combating} train two networks to improve each other.
(2) \textbf{F-correction}~\cite{patrini2017making} estimates the noise transition matrix to correct the loss function. 
(3) \textbf{D2L}~\cite{ma2018dimensionality} learns to monitor the dimensionality of subspaces and adapts the loss functions accordingly. 
(4) \textbf{Iterative-CV}~\cite{chen2019understanding} iteratively increases the number of the selected samples to train the networks. 
(5) \textbf{MentorNet}~\cite{jiang2017mentornet} is an example-weighting method based on curriculum learning. \textbf{MentorMix}~\cite{jiang2020beyond} further combines the MentorNet with the Mixup~\cite{zhang2017mixup}. 
(6) \textbf{DivideMix}~\cite{li2020dividemix} addresses the corrupted labels in a semi-supervised learning fashion.
(7) \textbf{M-correction}~\cite{arazo2019unsupervised} estimates the probability of a sample being mislabelled and then corrects the loss accordingly.

\subsubsection{Results on synthetic noisy labels}
Table~\ref{tab:cifar_mwn} shows the results on the CIFAR-10 and CIFAR-100 datasets with symmetric label noises. 
For the compared methods, we directly cite the reported numbers in their papers except for MW-Net~\cite{shu2019meta} and L2R~\cite{ren2018learning} where we report the reproduced results. For our method, we report the average and standard deviation of over three training trials using different random seeds.
The gains over baseline methods are statistically significant at the p-value level of 0.05, according to the one-tailed t-test.
These results illustrate the effectiveness of our method on the synthetic noisy labels.

\subsubsection{Results on realistic noisy labels}

Table~\ref{tab:webvision} shows the results on the WebVision dataset. As shown, our method consistently outperforms the baselines, achieving the best accuracy on the validation sets of WebVision and ImageNet. 
In particular, our method performs favorably against very recent methods such as MentorMix~\cite{jiang2020beyond} and DivideMix~\cite{li2020dividemix} in the top-1 accuracy.

Table~\ref{tab:cnwl} shows the results on the CNWL dataset. We implement several strong baselines using their official codes released on the CIFAR-100 dataset, \eg, 
MentorMix~\cite{jiang2020beyond}.
Note that in order to use their implementation, we downsample the images of the CNWL Mini-ImageNet dataset from 84x84 to 32x32. This results in new benchmark numbers to compare our baseline methods, and supplements \cite{jiang2020beyond}'s results on 32x32 images.
More details are discussed in the Appendix E.
Table~\ref{tab:cnwl} shows that our method outperforms all baseline methods on the realistic web noisy labels. The result is notable because 1) it verifies our method on the challenging CNWL dataset; 2) it demonstrates our consistent improvement across all noise rates as a useful and robust feature since the underlying noise rate is often unknown in practice.

We also apply our method on the Clothing1M dataset, and achieve 74.4\% in top-1 accuracy without using extra clean data, which is comparable to recently published methods. 
The results on the above three datasets demonstrate that our method trained with a noisy validation set is effective for addressing the realistic noisy labels.

\subsection{Long-tailed recognition task}

\label{sec:exp_longtail}

In addition to the noisy training label problem, we also evaluate our method on the long-tailed recognition task.

\textbf{Datasets and implementation details.} Four imbalanced factors $\{100, 50, 20, 10\}$ are applied on the long-tailed CIFAR-10 and CIFAR-100~\cite{cui2019class}. 
The number of training samples for each class is randomly removed by
$n_i\mu^i$, where $i$ indicates the class index, $n_i$ is the original number of the training samples for the $i$-th class, and $\mu\in(0,1)$. The imbalanced factor is the ratio between the largest and the smallest class. Following~\cite{shu2019meta, Jamal_2020_CVPR}, we do not change the test set and select ten training images per class as the clean validation set. Our method is implemented on the MW-Net model~\cite{shu2019meta} with the ResNet-32 backbone~\cite{he2016deep}.

From Table~\ref{tab:long_tailed_cifar}, we find our method consistently outperforms previous meta-learning based methods~\cite{ren2018learning, shu2019meta, Jamal_2020_CVPR}. 
Moreover, our method accelerates the training of the meta-learning model MW-Net by 2.9 times.
It is noteworthy that even compared to the very recent approaches (\eg, improved L2R~\cite{Jamal_2020_CVPR}), our method still obtains a reasonable performance gain, which illustrates the effectiveness of our method on the long-tailed recognition task.

\section{Conclusion}
In this paper, we discuss a novel \methodname\space(\methodnameabbre) to efficiently approximate the meta gradients by a layer-wise meta gradient sampling fashion.
We empirically show that our method yields not only an accurate but also a low-variance approximation of the meta gradient. The experimental results demonstrate that \methodnameabbre\space is able to reduce two-thirds of the training time of the meta-learning methods, while achieving a better generalization performance. Our method yields the state-of-the-art performance to address the noisy label problem, and obtains competitive performance on the long-tailed recognition task.

We find meta-model training is considerably influenced by the quantity and quality of the pseudo-clean label set. Future research in this area may include improving the robustness on limited validation data or low-quality pseudo validation data, in addition to further closing the gap in training time.

\clearpage

{\small
\balance
\bibliographystyle{ieee_fullname}
\bibliography{egbib}

\begin{thebibliography}{10}\itemsep=-1pt

\bibitem{arazo2019unsupervised}
Eric Arazo, Diego Ortego, Paul Albert, Noel~E O'Connor, and Kevin McGuinness.
\newblock Unsupervised label noise modeling and loss correction.
\newblock In {\em ICML}, 2019.

\bibitem{cao2019learning}
Kaidi Cao, Colin Wei, Adrien Gaidon, Nikos Arechiga, and Tengyu Ma.
\newblock Learning imbalanced datasets with label-distribution-aware margin
  loss.
\newblock In {\em NeurIPS}, 2019.

\bibitem{chawla2002smote}
Nitesh~V Chawla, Kevin~W Bowyer, Lawrence~O Hall, and W~Philip Kegelmeyer.
\newblock Smote: synthetic minority over-sampling technique.
\newblock {\em Journal of artificial intelligence research}, 16:321--357, 2002.

\bibitem{chen2019understanding}
Pengfei Chen, Benben Liao, Guangyong Chen, and Shengyu Zhang.
\newblock Understanding and utilizing deep neural networks trained with noisy
  labels.
\newblock In {\em ICML}, 2019.

\bibitem{cheng2020advaug}
Yong Cheng, Lu Jiang, Wolfgang Macherey, and Jacob Eisenstein.
\newblock Advaug: Robust adversarial augmentation for neural machine
  translation.
\newblock In {\em ACL}, 2020.

\bibitem{cui2019class}
Yin Cui, Menglin Jia, Tsung-Yi Lin, Yang Song, and Serge Belongie.
\newblock Class-balanced loss based on effective number of samples.
\newblock In {\em CVPR}, 2019.

\bibitem{drummond2003c4}
Chris Drummond, Robert~C Holte, et~al.
\newblock C4. 5, class imbalance, and cost sensitivity: why under-sampling
  beats over-sampling.
\newblock In {\em Workshop on learning from imbalanced datasets II}, volume~11,
  pages 1--8. Citeseer, 2003.

\bibitem{goldberger2016training}
Jacob Goldberger and Ehud Ben-Reuven.
\newblock Training deep neural-networks using a noise adaptation layer.
\newblock In {\em ICLR}, 2017.

\bibitem{han2018co}
Bo Han, Quanming Yao, Xingrui Yu, Gang Niu, Miao Xu, Weihua Hu, Ivor Tsang, and
  Masashi Sugiyama.
\newblock Co-teaching: Robust training of deep neural networks with extremely
  noisy labels.
\newblock In {\em NeurIPS}, 2018.

\bibitem{han2005borderline}
Hui Han, Wen-Yuan Wang, and Bing-Huan Mao.
\newblock Borderline-smote: a new over-sampling method in imbalanced data sets
  learning.
\newblock In {\em International conference on intelligent computing}. Springer,
  2005.

\bibitem{he2009learning}
Haibo He and Edwardo~A Garcia.
\newblock Learning from imbalanced data.
\newblock {\em IEEE Transactions on knowledge and data engineering},
  21(9):1263--1284, 2009.

\bibitem{he2017mask}
Kaiming He, Georgia Gkioxari, Piotr Doll{\'a}r, and Ross Girshick.
\newblock Mask r-cnn.
\newblock In {\em ICCV}, 2017.

\bibitem{he2016deep}
Kaiming He, Xiangyu Zhang, Shaoqing Ren, and Jian Sun.
\newblock Deep residual learning for image recognition.
\newblock In {\em CVPR}, 2016.

\bibitem{he2016identity}
Kaiming He, Xiangyu Zhang, Shaoqing Ren, and Jian Sun.
\newblock Identity mappings in deep residual networks.
\newblock In {\em ECCV}, 2016.

\bibitem{hendrycks2018using}
Dan Hendrycks, Mantas Mazeika, Duncan Wilson, and Kevin Gimpel.
\newblock Using trusted data to train deep networks on labels corrupted by
  severe noise.
\newblock In {\em NeurIPS}, 2018.

\bibitem{Jamal_2020_CVPR}
Muhammad~Abdullah Jamal, Matthew Brown, Ming-Hsuan Yang, Liqiang Wang, and
  Boqing Gong.
\newblock Rethinking class-balanced methods for long-tailed visual recognition
  from a domain adaptation perspective.
\newblock In {\em CVPR}, 2020.

\bibitem{jang2016categorical}
Eric Jang, Shixiang Gu, and Ben Poole.
\newblock Categorical reparameterization with gumbel-softmax.
\newblock In {\em ICLR}, 2017.

\bibitem{jiang2020beyond}
Lu Jiang, Di Huang, Mason Liu, and Weilong Yang.
\newblock Beyond synthetic noise: Deep learning on controlled noisy labels.
\newblock In {\em ICML}, 2020.

\bibitem{jiang2015self}
Lu Jiang, Deyu Meng, Qian Zhao, Shiguang Shan, and Alexander Hauptmann.
\newblock Self-paced curriculum learning.
\newblock In {\em AAAI}, 2015.

\bibitem{jiang2017mentornet}
Lu Jiang, Zhengyuan Zhou, Thomas Leung, Li-Jia Li, and Li Fei-Fei.
\newblock Mentornet: Learning data-driven curriculum for very deep neural
  networks on corrupted labels.
\newblock In {\em ICML}, 2018.

\bibitem{khan2017cost}
Salman~H Khan, Munawar Hayat, Mohammed Bennamoun, Ferdous~A Sohel, and Roberto
  Togneri.
\newblock Cost-sensitive learning of deep feature representations from
  imbalanced data.
\newblock {\em IEEE transactions on neural networks and learning systems},
  29(8):3573--3587, 2017.

\bibitem{krizhevsky2009learning}
Alex Krizhevsky, Geoffrey Hinton, et~al.
\newblock Learning multiple layers of features from tiny images.
\newblock 2009.

\bibitem{krizhevsky2012imagenet}
Alex Krizhevsky, Ilya Sutskever, and Geoffrey~E Hinton.
\newblock Imagenet classification with deep convolutional neural networks.
\newblock In {\em NeurIPS}, 2012.

\bibitem{li2020dividemix}
Junnan Li, Richard Socher, and Steven~C.H. Hoi.
\newblock Dividemix: Learning with noisy labels as semi-supervised learning.
\newblock In {\em ICLR}, 2020.

\bibitem{li2019learning}
Junnan Li, Yongkang Wong, Qi Zhao, and Mohan~S Kankanhalli.
\newblock Learning to learn from noisy labeled data.
\newblock In {\em CVPR}, 2019.

\bibitem{li2017webvision}
Wen Li, Limin Wang, Wei Li, Eirikur Agustsson, and Luc Van~Gool.
\newblock Webvision database: Visual learning and understanding from web data.
\newblock {\em arXiv preprint arXiv:1708.02862}, 2017.

\bibitem{liang2020simaug}
Junwei Liang, Lu Jiang, and Alexander Hauptmann.
\newblock Simaug: Learning robust representations from simulation for
  trajectory prediction.
\newblock In {\em ECCV}, 2020.

\bibitem{lin2017focal}
Tsung-Yi Lin, Priya Goyal, Ross Girshick, Kaiming He, and Piotr Doll{\'a}r.
\newblock Focal loss for dense object detection.
\newblock In {\em ICCV}, 2017.

\bibitem{liu2016ssd}
Wei Liu, Dragomir Anguelov, Dumitru Erhan, Christian Szegedy, Scott Reed,
  Cheng-Yang Fu, and Alexander~C Berg.
\newblock Ssd: Single shot multibox detector.
\newblock In {\em ECCV}, 2016.

\bibitem{liu2019large}
Ziwei Liu, Zhongqi Miao, Xiaohang Zhan, Jiayun Wang, Boqing Gong, and Stella~X
  Yu.
\newblock Large-scale long-tailed recognition in an open world.
\newblock In {\em CVPR}, 2019.

\bibitem{ma2018dimensionality}
Xingjun Ma, Yisen Wang, Michael~E Houle, Shuo Zhou, Sarah~M Erfani, Shu-Tao
  Xia, Sudanthi Wijewickrema, and James Bailey.
\newblock Dimensionality-driven learning with noisy labels.
\newblock In {\em ICML}, 2018.

\bibitem{mahajan2018exploring}
Dhruv Mahajan, Ross Girshick, Vignesh Ramanathan, Kaiming He, Manohar Paluri,
  Yixuan Li, Ashwin Bharambe, and Laurens van~der Maaten.
\newblock Exploring the limits of weakly supervised pretraining.
\newblock In {\em ECCV}, 2018.

\bibitem{malach2017decoupling}
Eran Malach and Shai Shalev-Shwartz.
\newblock Decoupling" when to update" from" how to update".
\newblock In {\em NeurIPS}, 2017.

\bibitem{miller2017reducing}
Andrew Miller, Nick Foti, Alexander D'Amour, and Ryan~P Adams.
\newblock Reducing reparameterization gradient variance.
\newblock In {\em NeurIPS}, 2017.

\bibitem{neelakantan2015adding}
Arvind Neelakantan, Luke Vilnis, Quoc~V Le, Ilya Sutskever, Lukasz Kaiser,
  Karol Kurach, and James Martens.
\newblock Adding gradient noise improves learning for very deep networks.
\newblock {\em arXiv preprint arXiv:1511.06807}, 2015.

\bibitem{northcutt2019confident}
Curtis~G Northcutt, Lu Jiang, and Isaac~L Chuang.
\newblock Confident learning: Estimating uncertainty in dataset labels.
\newblock {\em Journal of Artificial Intelligence Research}, 2021.

\bibitem{patrini2017making}
Giorgio Patrini, Alessandro Rozza, Aditya Krishna~Menon, Richard Nock, and
  Lizhen Qu.
\newblock Making deep neural networks robust to label noise: A loss correction
  approach.
\newblock In {\em CVPR}, 2017.

\bibitem{pleiss2020identifying}
Geoff Pleiss, Tianyi Zhang, Ethan~R Elenberg, and Kilian~Q Weinberger.
\newblock Identifying mislabeled data using the area under the margin ranking.
\newblock {\em arXiv preprint arXiv:2001.10528}, 2020.

\bibitem{redmon2016you}
Joseph Redmon, Santosh Divvala, Ross Girshick, and Ali Farhadi.
\newblock You only look once: Unified, real-time object detection.
\newblock In {\em CVPR}, 2016.

\bibitem{ren2018learning}
Mengye Ren, Wenyuan Zeng, Bin Yang, and Raquel Urtasun.
\newblock Learning to reweight examples for robust deep learning.
\newblock In {\em ICML}, 2018.

\bibitem{ren2016faster}
Shaoqing Ren, Kaiming He, Ross Girshick, and Jian Sun.
\newblock Faster r-cnn: Towards real-time object detection with region proposal
  networks.
\newblock {\em IEEE transactions on pattern analysis and machine intelligence},
  39(6):1137--1149, 2016.

\bibitem{shen2016relay}
Li Shen, Zhouchen Lin, and Qingming Huang.
\newblock Relay backpropagation for effective learning of deep convolutional
  neural networks.
\newblock In {\em ECCV}, 2016.

\bibitem{shu2019meta}
Jun Shu, Qi Xie, Lixuan Yi, Qian Zhao, Sanping Zhou, Zongben Xu, and Deyu Meng.
\newblock Meta-weight-net: Learning an explicit mapping for sample weighting.
\newblock In {\em NeurIPS}, 2019.

\bibitem{shu2020meta}
Jun Shu, Qian Zhao, Zengben Xu, and Deyu Meng.
\newblock Meta transition adaptation for robust deep learning with noisy
  labels.
\newblock {\em arXiv preprint arXiv:2006.05697}, 2020.

\bibitem{su2017pose}
Chi Su, Jianing Li, Shiliang Zhang, Junliang Xing, Wen Gao, and Qi Tian.
\newblock Pose-driven deep convolutional model for person re-identification.
\newblock In {\em ICCV}, pages 3960--3969, 2017.

\bibitem{szegedy2016inception}
Christian Szegedy, Sergey Ioffe, Vincent Vanhoucke, and Alex Alemi.
\newblock Inception-v4, inception-resnet and the impact of residual connections
  on learning.
\newblock In {\em AAAI}, 2016.

\bibitem{tanaka2018joint}
Daiki Tanaka, Daiki Ikami, Toshihiko Yamasaki, and Kiyoharu Aizawa.
\newblock Joint optimization framework for learning with noisy labels.
\newblock In {\em CVPR}, 2018.

\bibitem{vahdat2017toward}
Arash Vahdat.
\newblock Toward robustness against label noise in training deep discriminative
  neural networks.
\newblock In {\em NeurIPS}, 2017.

\bibitem{vinyals2016matching}
Oriol Vinyals, Charles Blundell, Timothy Lillicrap, Daan Wierstra, et~al.
\newblock Matching networks for one shot learning.
\newblock In {\em NeurIPS}, 2016.

\bibitem{vyas2020learning}
Nidhi Vyas, Shreyas Saxena, and Thomas Voice.
\newblock Learning soft labels via meta learning.
\newblock {\em arXiv preprint arXiv:2009.09496}, 2020.

\bibitem{wang2020training}
Zhen Wang, Guosheng Hu, and Qinghua Hu.
\newblock Training noise-robust deep neural networks via meta-learning.
\newblock In {\em CVPR}, 2020.

\bibitem{wei2020combating}
Hongxin Wei, Lei Feng, Xiangyu Chen, and Bo An.
\newblock Combating noisy labels by agreement: A joint training method with
  co-regularization.
\newblock In {\em CVPR}, 2020.

\bibitem{xia2019anchor}
Xiaobo Xia, Tongliang Liu, Nannan Wang, Bo Han, Chen Gong, Gang Niu, and
  Masashi Sugiyama.
\newblock Are anchor points really indispensable in label-noise learning?
\newblock In {\em NeurIPS}, 2019.

\bibitem{xiao2015learning}
Tong Xiao, Tian Xia, Yi Yang, Chang Huang, and Xiaogang Wang.
\newblock Learning from massive noisy labeled data for image classification.
\newblock In {\em CVPR}, 2015.

\bibitem{yang2019snapshot}
Chenglin Yang, Lingxi Xie, Chi Su, and Alan~L Yuille.
\newblock Snapshot distillation: Teacher-student optimization in one
  generation.
\newblock In {\em CVPR}, 2019.

\bibitem{yi2019probabilistic}
Kun Yi and Jianxin Wu.
\newblock Probabilistic end-to-end noise correction for learning with noisy
  labels.
\newblock In {\em CVPR}, 2019.

\bibitem{yin2019feature}
Xi Yin, Xiang Yu, Kihyuk Sohn, Xiaoming Liu, and Manmohan Chandraker.
\newblock Feature transfer learning for face recognition with under-represented
  data.
\newblock In {\em CVPR}, 2019.

\bibitem{yu2019does}
Xingrui Yu, Bo Han, Jiangchao Yao, Gang Niu, Ivor~W Tsang, and Masashi
  Sugiyama.
\newblock How does disagreement help generalization against label corruption?
\newblock In {\em ICML}, 2019.

\bibitem{zagoruyko2016wide}
Sergey Zagoruyko and Nikos Komodakis.
\newblock Wide residual networks.
\newblock In {\em BMCV}, 2016.

\bibitem{zhang2016understanding}
Chiyuan Zhang, Samy Bengio, Moritz Hardt, Benjamin Recht, and Oriol Vinyals.
\newblock Understanding deep learning requires rethinking generalization.
\newblock In {\em ICLR}, 2017.

\bibitem{zhang2017mixup}
Hongyi Zhang, Moustapha Cisse, Yann~N Dauphin, and David Lopez-Paz.
\newblock mixup: Beyond empirical risk minimization.
\newblock In {\em ICLR}, 2018.

\bibitem{zhang2020distilling}
Zizhao Zhang, Han Zhang, Sercan~O Arik, Honglak Lee, and Tomas Pfister.
\newblock Distilling effective supervision from severe label noise.
\newblock In {\em CVPR}, 2020.

\bibitem{zhou2020bbn}
Boyan Zhou, Quan Cui, Xiu-Shen Wei, and Zhao-Min Chen.
\newblock Bbn: Bilateral-branch network with cumulative learning for
  long-tailed visual recognition.
\newblock In {\em CVPR}, 2020.

\bibitem{zhu2020inflated}
Linchao Zhu and Yi Yang.
\newblock Inflated episodic memory with region self-attention for long-tailed
  visual recognition.
\newblock In {\em CVPR}, 2020.

\end{thebibliography}
\balance
}

\end{document}